\documentclass[runningheads]{llncs}

\usepackage[mobile]{eccv}

\usepackage{eccvabbrv}

\usepackage{graphicx}
\usepackage{booktabs}
\usepackage{colortbl}
\usepackage{adjustbox}

\usepackage[accsupp]{axessibility}  

\usepackage[ruled,linesnumbered]{algorithm2e}

\usepackage{hyperref}

\usepackage{orcidlink}

\begin{document}

\title{Diffusion Models are Geometry Critics:\\ Single Image 3D Editing Using Pre-Trained Diffusion Priors} 

\titlerunning{Single Image 3D Editing Using Pre-Trained
Diffusion Priors}

\author{Ruicheng Wang\inst{1}\thanks{Work done during internship at Microsoft Rsearch Asia.}\orcidlink{0000-0003-3082-0512} \and
Jianfeng Xiang\inst{2\!\!\ \star}\orcidlink{0000-0002-7380-2480} \and
Jiaolong Yang\inst{3}\orcidlink{0000-0002-7314-6567} \and
Xin Tong\inst{3}\orcidlink{0000-0001-8788-2453}}

\authorrunning{R.~Wang, J.~Xiang et al.}

\institute{University of Science and Technology of China \and Tsinghua University\and Microsoft Research Asia}

\maketitle

\begin{abstract}
  We propose a novel image editing technique that enables 3D manipulations on single images, such as object rotation and translation.
  Existing 3D-aware image editing approaches typically rely on synthetic multi-view datasets for training specialized models, thus constraining their effectiveness on open-domain images featuring significantly more varied layouts and styles. 
  In contrast, our method directly leverages powerful image diffusion models trained on a broad spectrum of text-image pairs and thus retain their exceptional generalization abilities.
  This objective is realized through the development of an iterative novel view synthesis and geometry alignment algorithm. 
  The algorithm harnesses diffusion models for dual purposes: they provide appearance prior by predicting novel views of the selected object using estimated depth maps, and they act as a geometry critic by correcting misalignments in 3D shapes across the sampled views.
  Our method can generate high-quality 3D-aware image edits with large viewpoint transformations and high appearance and shape consistency with the input image, pushing the boundaries of what is possible with single-image 3D-aware editing. Project webpage: \url{https://wangrc.site/Diff3DEdit/}
\end{abstract}

\section{Introduction}
\label{sec:intro}

On a typical morning, an artist opens the 3D creation software and picks up where they left off the previous day. The 3D scene therein depicts an alley with stone pavement extending into the distance, flanked by old-style houses. As they view the rendered pictures, a misplaced wooden handcart situated right in the middle of the alley disrupts the scene's harmony. The artist selects the handcart, moves it to a more suitable location at the side, rotates it to align its handle against the wall, and scales it down for a natural fit with the adjacent haystack. Running the rendering once more, the pictures are now more visually pleasing. This task, although seemingly simple, requires sophisticated 3D software and 3D scenes that take considerable time to construct.

\begin{figure}[t!]
  \includegraphics[width=\textwidth]{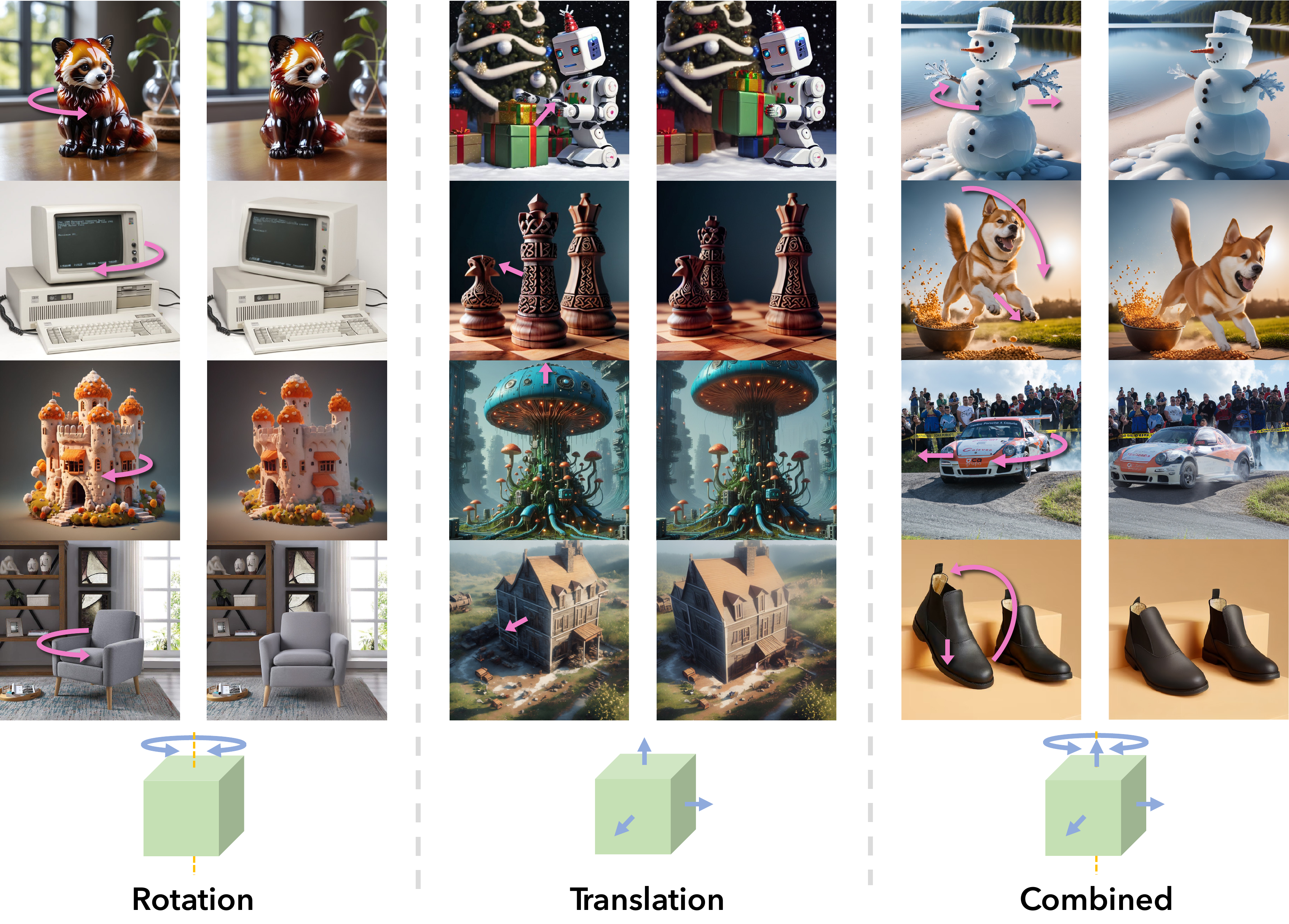}
  \caption{3D-aware image editing results of our proposed method. Our method enables 3D manipulations of objects with consistent appearance, plausible layout, and harmonious composition including occlusion (e.g., the first two examples of translation editing), by using pre-trained diffusion models. (\textbf{Best viewed with zoom-in})
  }
  \label{fig:teaser}
\end{figure}

Nowadays, thanks to the tremendous advancements in large-scale text-to-image generation models~\cite{saharia2022photorealistic,rombach2022high,chen2023pixartalpha}, the creation of delicate images has been revolutionized without the necessity of photography or 3D software. However, subsequent manipulation of these images in 3D space akin to editing with 3D software have not yet been satisfactorily achieved, leaving single-image 3D-aware editing a missing piece of jigsaw for AI-based image editing tool chains.





Diving into the former literature of generative image editing realm, many methods have been proposed with different generative modeling frameworks. Some of them have achieved 3D-aware editing to varied extents. Under the GAN paradigm~\cite{goodfellow2014generative}, a group of work~\cite{abdal2021styleflow,shen2020interpreting,harkonen2020ganspace,pan20202d,roich2022pivotal} investigated GAN model's compact and semantic-rich latent spaces 
and developed different techniques to find latent codes or directions reflecting the 3D-aware manipulations of the input images. Another group of works resorted to the training of disentangled GANs with 3D-related attributes directly as input~\cite{tewari2020stylerig,deng2020disentangled}, hence enabling explicit 3D manipulation with final models. 
However, due to the relatively limited modeling capacity of GANs, the applicability of these works is significantly constrained. 


Since diffusion models~\cite{sohl2015deep,ho2020denoising,song2020score} were proposed, they soon gain attention for their strong generative modeling capabilities across various domains. This triggered the development of diffusion-based image editing methods. Among them, a few works~\cite{michel2023object,yuan2023customnet} also aimed at 3D-aware image editing. To do so, they constructed training image pairs corresponding to certain 3D manipulations with synthetic data from 3D model collections and trained specialized diffusion models. This inevitably brought a negative impact to their effectiveness on open-domain images that feature significantly more varied layouts and styles. Another stream of works worth noticing is generative novel view synthesis~\cite{watson2022novel,chan2023genvs,liu2023zero}, which also models the 3D changes of given images like 3D-aware editing. 
However, they still face generalization challenges as the models are trained with either limited real-world video sequences or synthetic 3D renderings.
In a related yet slightly different realm, a number of works~\cite{xu2023neurallift,deng2023nerdi,Tang_2023_ICCV} lift images to 3D by distilling knowledge from pre-trained 2D diffusion models. This is typically achieved by optimizing a 3D radiance field with Score Distillation~\cite{poole2022dreamfusion} objectives, given single-view images as conditions. Though they enjoy the generalization ability of 2D diffusion models, the time-consuming radiance field optimization takes several hours to converge, and the results may suffer from visual quality degradations.




In this work, we aim to find an efficient and generalized way to solve the single-image 3D-aware  editing problem. 
A tuning-free approach is proposed using pre-trained image diffusion models. Our method does not require additional training on multiview datasets -- which are usually limited or biased in distribution -- except for a simple and efficient single-image LoRA training. It directly leverages powerful large-scale T2I diffusion models~\cite{rombach2022high}, 
preserving their exceptional generalization abilities and high image quality.

The underpinning of our method is a novel depth-warp-assisted iterative algorithm for image generation and refinement, which is guided by diffusion priors. The central insight of our approach is that diffusion models are capable of serving a dual purpose: they not only provide priors for high-fidelity texture completion of the target-view image but also act as a geometry critic, correcting distortions caused by inaccurate depth estimation. By utilizing diffusion-based image undistortion, we further refine the depth map of the original image and align it with the target-view depth. This creates a feedback loop that facilitates iterative view synthesis and continuous improvement of the shape representation.

Our method enjoys 1) the texture consistency benefited from image warping, 2) the shape consistency achieved by the iterative shape improvement loop with the guidance of diffusion models, and 3) the high overall image quality and visual harmony afforded by the diffusion priors as well. It does not require model training/fine-tuning on multiview dataset and operates effectively on open-domain images.

\textbf{The contributions of this work} are summarized below:
\begin{itemize}
    \item We propose a novel depth-assisted single-image 3D-aware image editing method that leverages pre-trained large image diffusion models and works on open-domain images.
    \item We introduce a closed-loop iterative algorithm for harnessing the geometric priors inherent in image diffusion models, with the key insight being that these models can rectify geometric distortions and hence can be leveraged to enhance geometry reasoning.
    \item We achieve visually pleasing and geometrically consistent 3D editing results that cannot be achieved by previous methods, and an interactive system is implemented for user-friendly 3D-aware image editing.
\end{itemize}

\section{Related Work}

\paragraph{Diffusion models.}


Diffusion models~\cite{sohl2015deep} are grounded in a solid theoretical framework and are later complemented by well-designed network architectures~\cite{ho2020denoising, song2020score}, rendering them highly effective for generative image modeling tasks. Subsequent advancements in diffusion-based methods~\cite{nichol2021improved, ho2022cascaded, dhariwal2021diffusion, ho2022classifier, karras2022elucidating, bao2023all, peebles2023scalable} have established diffusion models (DMs) as the new state-of-the-art models, eclipsing GANs in certain image generation tasks.

Further, image diffusion models trained on large open-domain images, such as Imagen~\cite{saharia2022photorealistic}, Stable Diffusion~\cite{rombach2022high}, and PixArt-$\alpha$~\cite{chen2023pixartalpha}, have showcased the remarkable scalability and generalizability of diffusion models. This has encouraged the application of these models in subsequent research to leverage their broad understanding of image content. In this work, we utilize these pre-trained, large-scale diffusion models as priors to achieve 3D-aware editing on single images.

\paragraph{Generative image editing and 3D-aware image editing.}
In the realm of generative image editing, many previous image editing methods are based on GANs~\cite{goodfellow2014generative,karras2019style,karras2020analyzing}, such as \cite{abdal2021styleflow,
shen2020interpreting,harkonen2020ganspace,roich2022pivotal,pan2023drag}. Among them, 3D-aware image editing~\cite{pan20202d,tewari2020stylerig,deng2020disentangled,zhou2020rotate} aims to manipulate the image content in a manner that reflects certain changes in 3d space, such as viewpoint shifts. However, given the limited capacity of GANs, the effectiveness of these methods is limited to specific categories of images.

More recently, as diffusion models surpass GANs for their remarkable generative modeling capacity, there has been a surge of diffusion-based image editing methods~\cite{brooks2023instructpix2pix,geng2023instructdiffusion,sheynin2023emu,chen2023anydoor,kawar2023imagic,song2023objectstitch}. 
Among these, tuning-free approaches~\cite{hertz2022prompt,mokady2022null,meng2022sdedit,tumanyan2023plug,cao_2023_masactrl,parmar2023zero,epstein2023selfguidance,mou2023dragondiffusion,shi2023dragdiffusion} are particularly noteworthy. These methods utilize pre-trained diffusion models without additional fine-tuning, thereby preserving the models' inherent ability to generalize.
For 3D-aware image editing, existing diffusion-based methods~\cite{yuan2023customnet,michel2023object} opt to use specialized models trained on synthetic image pairs. However, the domain gap between these synthesized images and real-world images often limits their applicability.

Our proposed method achieves tuning-free 3D-aware editing with large-scale diffusion foundation models, and hence exhibits significant generalization ability across open-domain images.

\paragraph{Novel view synthesis.}
Novel view synthesis (NVS) aims to synthesize a target image with an arbitrary target camera pose from given source images and their camera poses. Traditional methods~\cite{yu2021pixelnerf,sajjadi2022scene,wang2021ibrnet,chen2021mvsnerf,han2022single} formulate novel view synthesis task as a regression problem, leading to blurry prediction when the target view deviates significantly from the source. On the other hand, recent diffusion-based approaches~\cite{liu2023zero,watson2022novel,chan2023genvs,ye2023consistent,long2023wonder3d,shi2023zero123++,xiang2023ivid} formulate it as a generation task and have achieved high synthesis quality for novel views. However, they typically rely on models trained on limited real-world multiview images multiple viewpoints or synthetic images from 3D models. Their generalization ability is thereby constrained. 
NVS can also be achieved by single image to 3D method~\cite{xu2023neurallift,deng2023nerdi,Tang_2023_ICCV} utilizing 2D diffusion model's prior knowledge with score distillation technique. These methods typically involve radiance fields optimization and hence are time-consuming and may suffer from quality degradation compared to the 2D diffusion models.

\begin{figure*}[t!]
	\centering
	\includegraphics[width=0.8\textwidth]{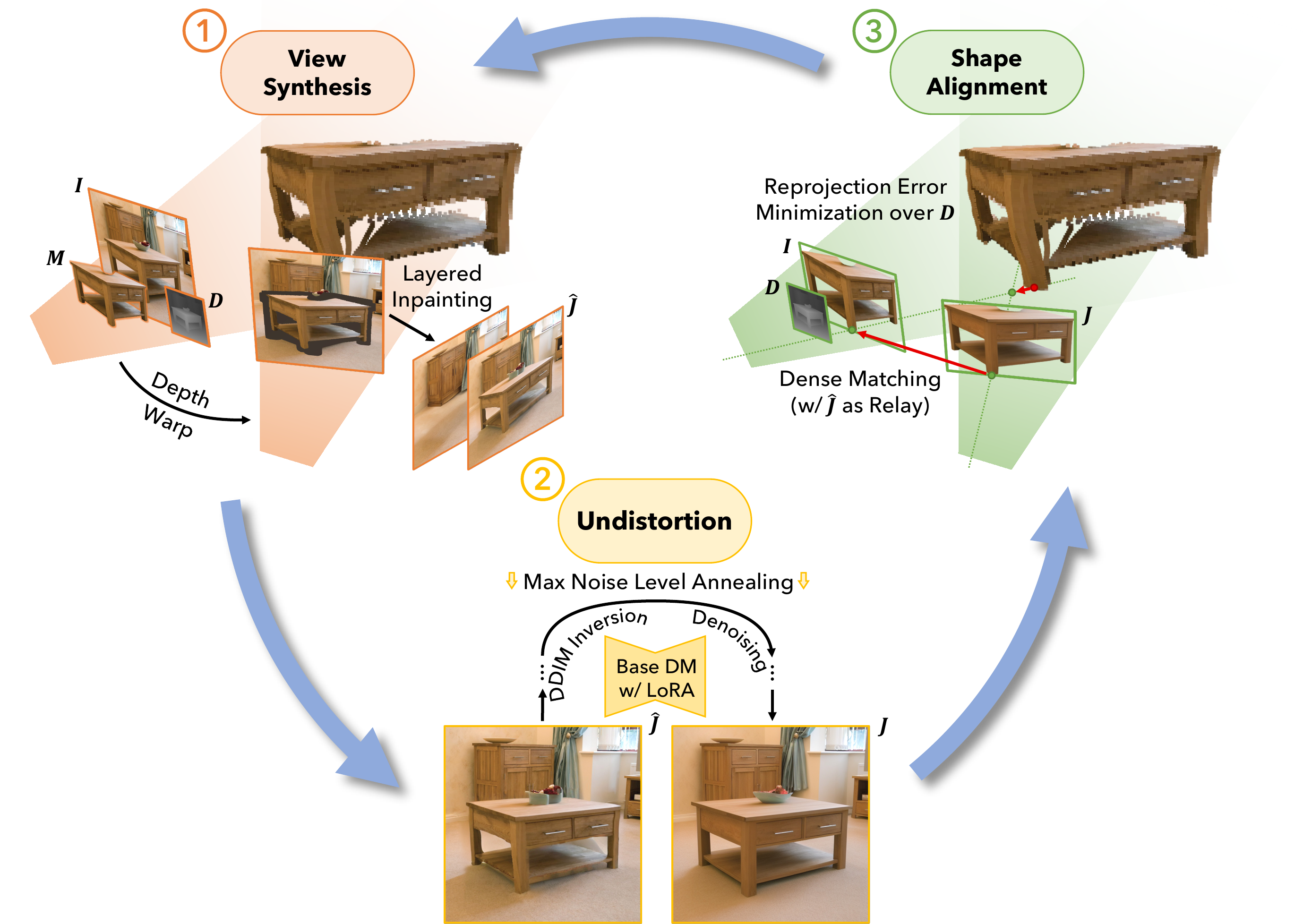}
	\caption{The overall pipeline. Our 3D-aware editing method iterates among three phases. 
 The \emph{view synthesis} phase generates the novel view of the selected object using depth-based warping and layered diffusion inpainting (initial depth map obtained by monocular depth estimation). The \emph{undistortion} phase rectifies the potential distortions on target-view image induced by inferior depth estimate. The \emph{shape alignment} phase aligns the object shape in the original input image to the undistorted target image by optimizing the depth map and minimizing dense image correspondences. After several iterations, this process yields plausible and consistent 3D editing results.}
	\label{fig:pipeline}
\end{figure*}

\section{Approach}


Our goal in this paper is to enable single image 3D-aware editing, such as 3D rotation and translation, applicable for open-domain images. We base our method on the powerful T2I diffusion models pre-trained on large-scale datasets and frozen in our framework.
The method iteratively operates in three phases, effectively utilizing appearance and geometric priors from diffusion models. As shown in Figure~\ref{fig:pipeline}, the \emph{view synthesis} phase (Sec.~\ref{sec:warp}) creates a new view of the target object by depth-based warping and diffusion-based image completion. The \emph{undistortion} phase (Sec.~\ref{sec:undistortion}) then corrects the warping-induced distortions with a given max noise level parameter. Lastly, in the \emph{shape alignment} phase (Sec.~\ref{sec:geometry}), a geometric solver adjusts the object's shape to match the new view through depth adjustment. Our overall algorithm (Sec.~\ref{sec:algorithm}) can produce high-fidelity 3D-aware edits by iterating these three phases with the max noise level parameter of the undistortion phase gradually decreasing.



\subsection{View Synthesis} \label{sec:warp}

In the first phase, our goal is to predict the novel view of the given object after applying instructed 3D transformations. Two requirements must be achieved for this goal: for the parts of the object that are still visible after transformation, they should be consistent with original image; for other parts that are originally occluded but visible in the target image, they should be generated and remain compatible with other image content.

We opt for depth-assisted image warping -- which lifts an image to a 3D mesh using depth and projects it onto image plane again after applying 3D transformations -- as our starting point. It maintains all texture details and strictly follows the given 3D transformation for the visible parts. Initially, we obtain a depth map using the off-the-shelf monocular depth estimator of \cite{bhat2023zoedepth}.
To fill in the holes after depth-based warping and obtain a complete image, a straightforward solution is to apply image inpainting. 
However, naive hole inpainting could generate undesirable layout that is incompatible to the contents viewed from source image. For example, parts known to be background could be mistakenly filled by extending the foreground object and vice versa. 

\begin{figure}[t]
	\centering
	\includegraphics[width=0.7\textwidth]{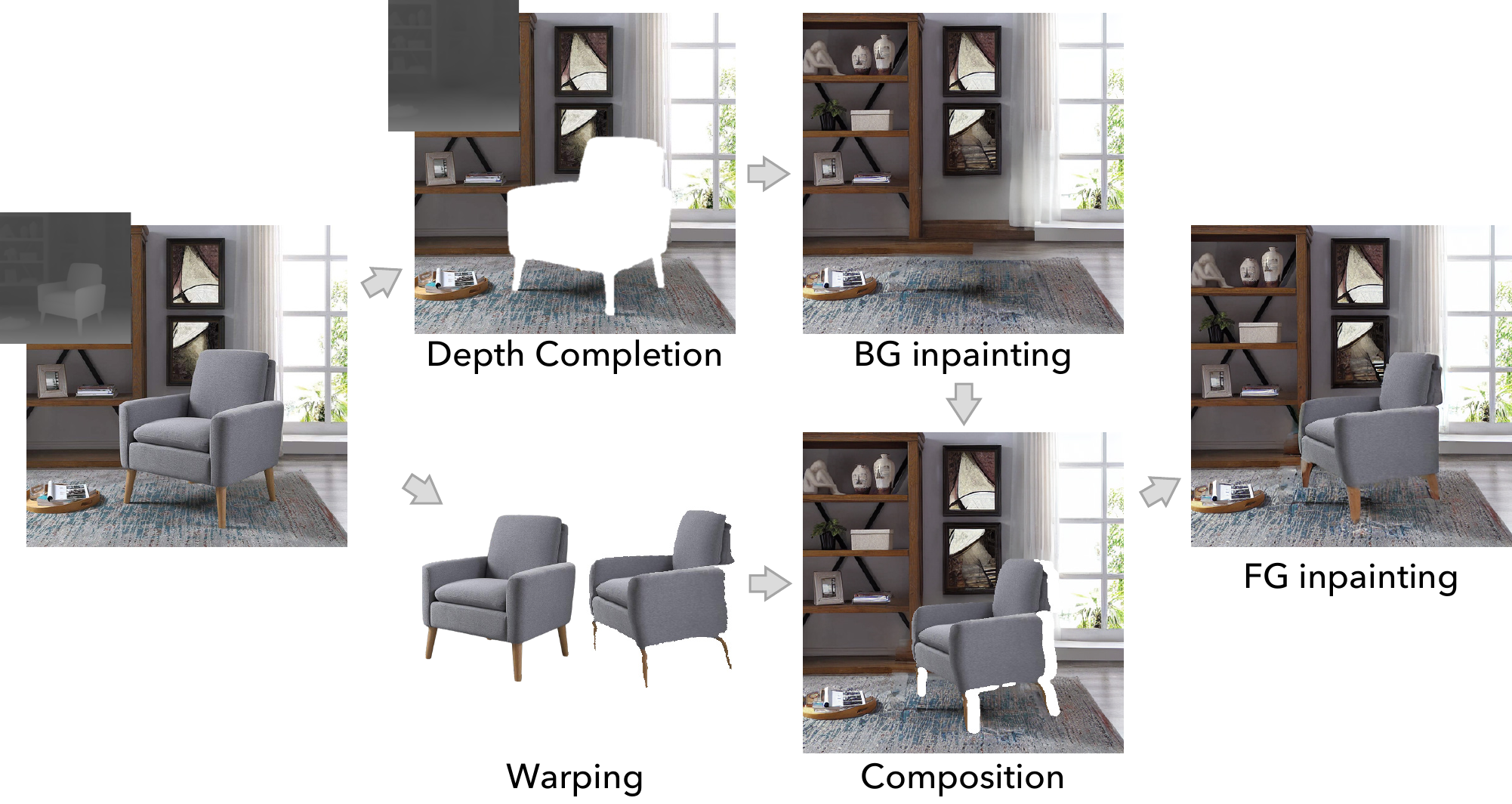}
	\caption{Illustration of our view synthesis stage with a layered, diffusion-based generative inpainting scheme.}
	\label{fig:view_synthesis}
\end{figure}

Therefore, we design a layered inpainting process tailored for our view synthesis task, which is illustrated in Figure~\ref{fig:view_synthesis}. Specifically, we first complete the background layer of the raw input image which is occluded by the selected foreground object. Naive inpainting cannot guarantee the filled contents are pure background (\emph{e.g.}, new random foreground objects could appear). To address this, we first inpaint the depth map of the input image by a simple heuristic algorithm (see the \emph{suppl. material} for more detail), and then inpaint the color content by combining the image inpainting DM and depth-guided ControlNet~\cite{zhang2023adding}.

To accommodate for the object's thickness along the Z-axis, we replicate the edge pixels of the foreground and deepen their depth values prior to 3D mesh construction. 
Upon transforming and reprojecting the 3D mesh onto the image plane, new ambiguous areas emerge, which may belong to either the foreground or background. We merge this warped image, including the uncertain regions, with the inpainted background, and employ generative inpainting once more.
This way, the resulting novel view synthesis meets our criteria, ensuring a cohesive and harmonious appearance with a compatible layout. More implementation details of the view synthesis process can be found in the \emph{suppl. material}. 



\subsection{Undistortion} \label{sec:undistortion}

\begin{figure*}[t]
	\centering
	\includegraphics[width=1\textwidth]{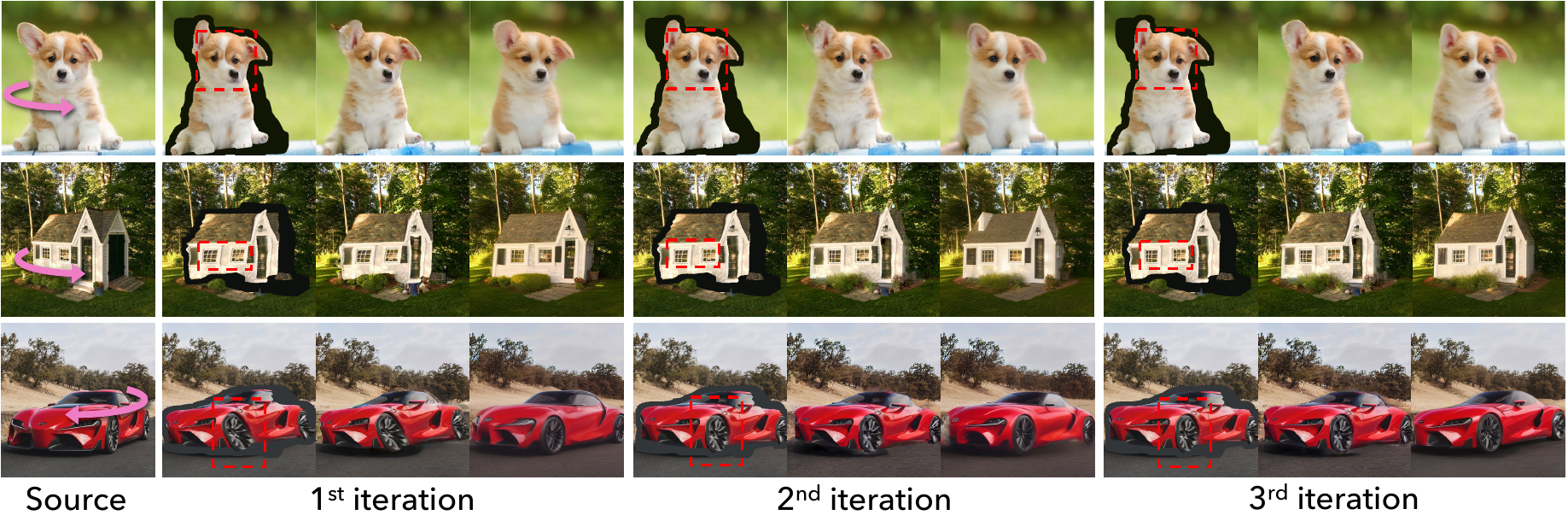}
	\caption{Intermediate results of our iterative algorithm. (Best viewed with zoom-in; see text in Sec. \ref{sec:visual} for detailed explainations)}
	\label{fig:vis}
\end{figure*}

The novel view synthesis procedure is vulnerable to poor depth quality. As shown in Figure~\ref{fig:vis}, the image can exhibit significant geometric distortions when subjected to 3D rotation. Even minor irregularities in the 3D shape may induce dramatic changes in layout when viewed from new angles. Unfortunately, even the most advanced monocular depth estimators today fall short of delivering flawless results for arbitrary images.



As such, we introduce an undistortion process which harnesses the inherent geometric priors of a diffusion model. This process is designed to minimize undesirable distortions (such as bent lines and stretched textures) meanwhile maintaining a consistent appearance to the extent possible. Our solution is inspired by SDEdit~\cite{meng2022sdedit}, which provides a simple yet effective editing method based on two key premises. First, diffusion models simulate the distribution of real images at various levels of noise. Second, the distributions of the real and conditioning images converge. Consequently, integrating the conditioning image into the real image distribution—by adding noise and then denoising with diffusion models—has proven to be an effective editing technique.

Our undistortion process adapts such a noising-denoising algorithm with two modifications to meet the unique requirements of our undistortion task, i.e., the preservation of consistent appearance. \emph{First}, we employ a low-rank adaptation (LoRA)~\cite{hu2021lora} to the attention layers of the diffusion model, trained with the input source image for dozens of steps, which takes only a few seconds. The LoRA provides global conditioning to guide the sampling process semantically. \emph{Second}, for fine-level appearance consistency, we modify the noising-denoising process used in SDEdit by replacing noise addition with DDIM inversion~\cite{song2020denoising}. This change is motivated by the fact that the earlier-step perturbation in the ODE numeric solving process will bring much higher difference in the final results, compared to later steps. 
Therefore, by restricting perturbations to the denoising phase and employing deterministic DDIM inversion for noising, we strike a balance that preserves appearance consistency while leveraging the diffusion model's capacity to correct the image. 
Implementation details are provided in \emph{suppl. material}.

The undistortion process features a \emph{max noise level} parameter $\sigma$ which controls the extent of image deviation allowed. 
A higher value gives the model greater freedom to alter the content, enabling it to correct larger distortions but potentially at the expense of increased texture detail inconsistency. Conversely, a lower value restricts content changes, suitable for correcting smaller distortions while preserving more texture detail. 
This parameter is gradually adjusted throughout the iterations of our whole algorithm.

\subsection{Shape Alignment} \label{sec:geometry}
Although the results after undistortion usually exhibit plausible geometry, the consistency between source images and the editing results are sacrificed, especially at the early iterations with suboptimal depth perception and substantial distortion corrections. 
To enhance this consistency, we shall aim for more minor distortion adjustments, which in turn depend on more accurate depth estimations. Therefore, we implement a shape alignment step that enforces the depth of object in the original image conforms to the distortion-corrected target-view image.

To achieve this goal, we build dense correspondences between the original image $\boldsymbol{I}$ and the distortion-corrected target-view image $\boldsymbol{J}$ and then optimize the depth map $\boldsymbol{D}$ of $\boldsymbol{I}$ via reprojection error minimization. Given that $\boldsymbol{I}$ and $\boldsymbol{J}$ may exhibit changes in viewpoint, forming direct correspondences between them can be less unreliable. To address this, we use the target-view image before undistortion, denoted as $\hat{\boldsymbol{J}}$, as a correspondence bridge. This is because $\boldsymbol{I}$ has readily available correspondences with $\hat{\boldsymbol{J}}$ from image warping, and $\hat{\boldsymbol{J}}$ and $\boldsymbol{J}$ are view-aligned thus facilitating more dependable correspondence matching. We employ the method of \cite{truong2021learning} to identify matches between $\hat{\boldsymbol{J}}$ and $\boldsymbol{J}$. 

With the correspondences between $\boldsymbol{I}$ and $\boldsymbol{J}$, we apply a geometry solver to optimize $\boldsymbol{D}$ by minimizing the reprojection error of these correspondences points along with a gradient regularization term. This optimization task is formulated as a least squares problem:
\begin{equation}
\min_{\boldsymbol{D}}\sum_{(\boldsymbol{x_I},\boldsymbol{x_J})\in\boldsymbol{\mathcal{C}}}\|
    \Pi(\boldsymbol{x_I}, \boldsymbol{D}, \boldsymbol{T})
    -\boldsymbol{x_J}\|_2^2+\lambda\mathrm{Reg}(\boldsymbol{D}),
\end{equation}
where $\boldsymbol{\mathcal{C}}$ denotes the correspondence set in which $\boldsymbol{x_I}$ and $\boldsymbol{x_J}$ represent corresponding points in $\boldsymbol{I}$ and $\boldsymbol{J}$, respectively. $\Pi$ represents the projection function from the source to the target image given 3D transformation $\boldsymbol{T}$; camera intrinsics omitted for brevity.
Solving this optimization problem yields an enhanced 3D shape of the object that is aligned with the undistorted image as closely as possible, thereby correcting shape errors gauged by the plausibility of the layout from the new viewpoint. Please refer to \emph{suppl. material} for more details.

\begin{algorithm}[t]
\SetKwFunction{EstimateDepth}{EstimateDepth}
\SetKwFunction{LoRATuning}{LoRATuning}
\SetKwFunction{SynView}{SynView}
\SetKwFunction{Undistort}{Undistort}
\SetKwFunction{AlignShape}{AlignShape}
\SetKwInOut{Input}{input}
\SetKwInOut{Output}{output}
\caption{The main algorithm}\label{alg:selfcritical}
\Input{source image $\boldsymbol{I}$, selection mask $\boldsymbol{M}$,\\ 3D transformation $\boldsymbol{T}$, resampling times $N_\mathrm{rs}$}
\Output{3D-aware edits $\boldsymbol{J}$}
$\LoRATuning(\boldsymbol I)$\\
$\boldsymbol{D} \leftarrow \EstimateDepth(\boldsymbol{I})$\\
\For{$i\leftarrow 0$ \KwTo $N_\mathrm{rs}$}{
    $\hat{\boldsymbol{J}}\leftarrow\SynView(\boldsymbol{I}, \boldsymbol{M}, \boldsymbol{D}, \boldsymbol{T})$\\
    $\boldsymbol{J}\leftarrow\Undistort(\hat{\boldsymbol{J}})$ with noise level $\sigma$\\
    $\boldsymbol{D}\leftarrow\AlignShape(\boldsymbol{J}, \hat{\boldsymbol{J}}, \boldsymbol{I}, \boldsymbol{D}, \boldsymbol{T})$\\
    $\sigma\leftarrow\sigma\!\downarrow$\\
}
\end{algorithm}

\subsection{The Overall Algorithm} \label{sec:algorithm}

With all necessary components in place, our comprehensive algorithm is outlined in Algorithm~\ref{alg:selfcritical}.  During the initialization phase, we estimate the depth map of the input image and tune the low-rank adaptation of the Stable Diffusion model~\cite{rombach2022high} with the  image. Then the \emph{view synthesis}, \emph{undistortion}, and \emph{shape alignment} will be applied in a row several times. 

During the iterative process, we \emph{progressively decrease the maximum noise level} parameter in the undistortion phase. The process yields a continuous reduction in geometric distortion and a gradual increase in consistency between the final output and the original image.




\section{Experiments}

We conduct a series of experiments to demonstrate the effectiveness of our method on the single-image 3D-aware editing task and compare it against previous methods. 

\paragraph{Implementation details.} 
We employ the \emph{Stable Diffusion v1.5} as our base diffusion model. The number of iterations performed in the overall algorithm is set at $3$, with the max noise level parameter $\sigma$ set to $(0.5, 0.4, 0.3)$, where $1$ signifies the start and $0$ represents the end of the diffusion reverse chain. The LoRA is tuned for 60 steps. The strengths of classifier-free guidance and the number of total sampling steps are set to $(4.0, 25)$ for the inpainting process and $(1.0, 50)$ for the undistortion process, respectively. 

\paragraph{Running time.}
Tested on a single NVIDIA Tesla A100 GPU, the running time of our method for a single 3D editing operation is about 20 seconds. An additional preparation phase involving LoRA tuning requires roughly 10 more seconds.

\paragraph{Benchmark.}
Our experiments utilize an image corpus that is  designed with high diversity and coverage. The benckmark dataset spans over 10 categories that are frequently encountered in everyday life, including \emph{animal}, \emph{appliance}, \emph{building}, \emph{food}, \emph{furniture}, \emph{personal items}, \emph{toys}, \emph{nature}, \emph{street}, and \emph{vehicle}.
It contains 80 images with $512\times512$ resolution that are manually collected from the Internet. The objects in the images are segmented by the SAM model~\cite{kirillov2023segany}. Each image is annotated with $1\!\sim\!4$ instructed 3D rotation and translation edits, forming 160 test cases in total. 


\begin{figure*}[t]
	\centering
	\includegraphics[width=1\textwidth]{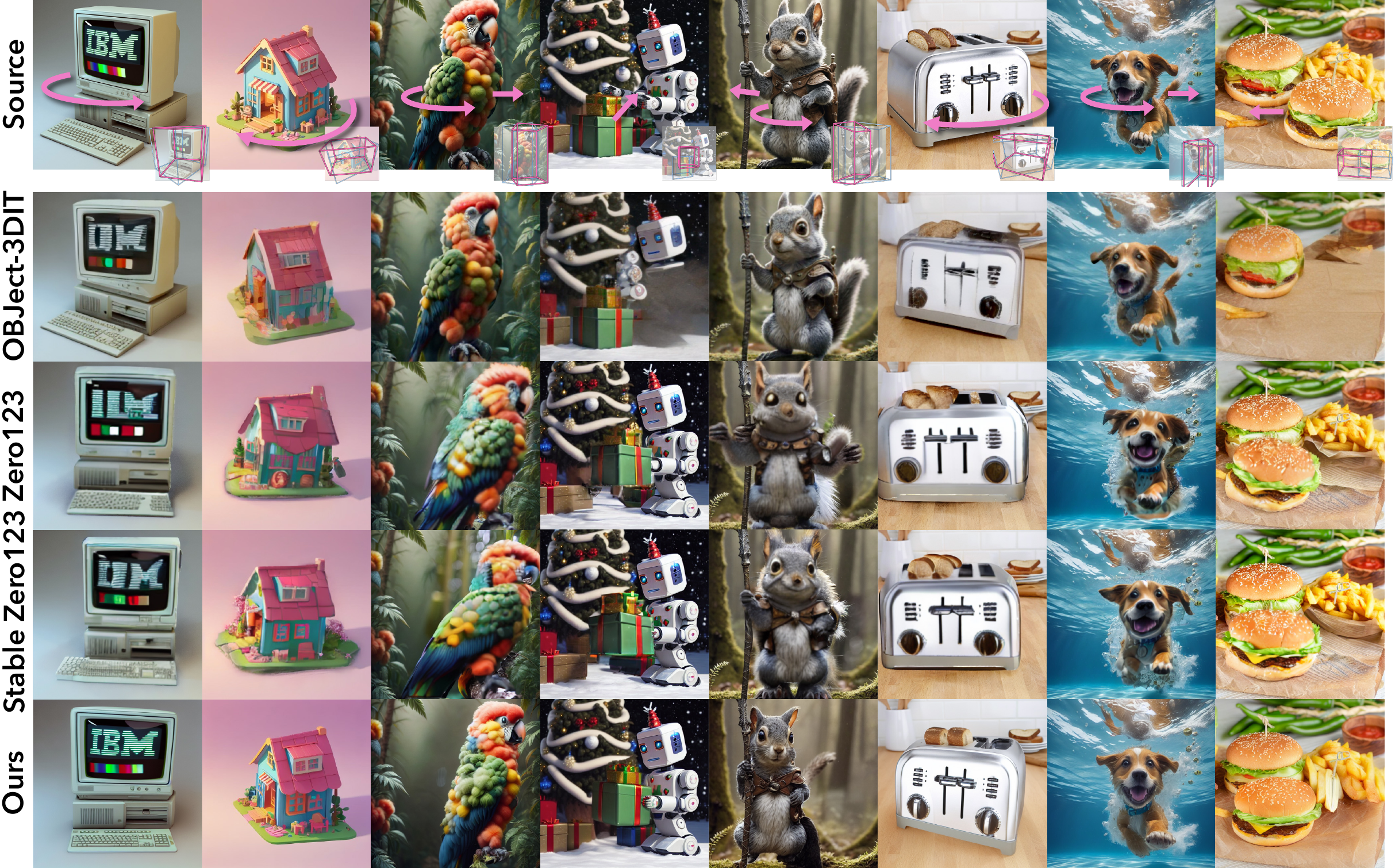}
	\caption{Visual comparison of different methods. The bottom-right figures in the first row depict the initial and target poses for editing. The results of Zero123 and Stable Zero123 are postprocessed by overlaying the generated objects onto the inpainted background.  (\textbf{Best viewed with zoom-in; see the \emph{suppl. material} for more results.})}
	\label{fig:results}
\end{figure*}

\subsection{Visual Results}\label{sec:visual}

\paragraph{Edited images.} Figure~\ref{fig:teaser}, Figure~\ref{fig:results} and Figure~\ref{fig:large_transformation} showcase some edited images produced by our method. The method is capable of generating high-quality 3D-aware image edits under large viewpoint transformations, while preserving the plausibility of the layout as well as the consistency with original images. 

\paragraph{Algorithm visualization.} 
To better illustrate the effectiveness of our proposed method, we provide visualizations of the intermediate results generated during the iterative process. The results are presented in Figure~\ref{fig:vis}. For the first iteration, the view synthesis phase first produces an imperfect sample of the new desired view through depth-based image warping (1st column) and layered diffusion inpainting (2nd column). Notable distortions in the layout structure, such as stretched window grids and tires, twisted facial features on dogs, and bent washer edges, are evident. These issues stem from inaccuracies in the monocular shape estimator. Subsequently, the new view is rectified to produce the images presented in the third column through the undistortion phase. The effect of distortion correction is obvious, thanks to the prior knowledge of geometric structures from the base diffusion model. In the third step, the shape is adjusted according to cross-view correspondences and 3D projection relations, and such an improvement leads to a less distorted warping in the second iteration.
The results are further improved in the second and third iterations, with gradually fewer geometry distortions and improved consistency to the original input image.

\subsection{Comparison with Previous Methods}

Several previous works have also investigated 3D-aware image editing or are closely related to this objective, including OBJect-3DIT\cite{michel2023object} and Zero123\cite{liu2023zero}. Note that unlike our method, these two approaches rely on \emph{training with synthetic renderings sourced from collections of 3D models}. Zero123 aims to generate novel views of a given object, which can be applied to the 3D-aware image editing task with rotation manipulation. We implemented a pipeline to enable the comparison to Zero123. Specifically, we rotate the foreground object with Zero123 and overlay its result onto the inpainted background. The 3D translation operation, which cannot be handled by Zero123, is approximated by 2D image translation and scaling after rotation. For a harmonious composition, an additional generative inpainting is applied to the surrounding region of the foreground object. For both OBJect-3DIT and Zero123, we use the pre-trained checkpoints provided by the authors' official implementations. We additionally include a checkpoint of Zero123 trained by Stability AI\footnote{https://huggingface.co/stabilityai/stable-zero123}, which demonstrated superior performance compared to original model. 


\paragraph{Qualitative comparison.}
Figure~\ref{fig:results} presents a comparison of editing results obtained with different methods. Upon visual inspection, it is apparent that OBJect-3DIT underperforms in terms of layout plausibility due to its limited training dataset. It fails to achieve 3D manipulation for a large portion of the test cases. Zero123 exhibits a degree of generalization capability; however, it still struggles with complex scenarios that deviate from its training examples, particularly with objects that have complex textures and features, such as furry animals.
In contrast, our method delivers plausible layouts across the majority of evaluation cases, indicating its advanced generalization capacity. 

In terms of image quality, the edits produced by OBJect-3DIT and Zero123 exhibit more visual artifacts and a tendency towards blurriness when compared to our method. This demonstrates the superiority of our approach which can leverage strong image generation models without further training/finetuning. The blurriness of OBJect-3DIT and Zero123 is also partially due to the lower training resolution of their base diffusion models ($256\times256$, \emph{vs.} ours $512\times512$).


Regarding appearance consistency, our method outperforms the other two, particularly in regions with highly detailed textures, such as text and fine lines. This is attributed to our method's reliance on view synthesis assisted by explicit warping, rather than an implicitly learned attention mechanism.


\paragraph{Quantitative comparison.}

We further perform a numerical evaluation of appearance consistency. For perceptual consistency, we employ the average cosine similarity between the CLIP~\cite{radford2021learning} image feature of the input and edited images. To measure more detailed local consistency, we use an off-the-shelf dense image matching method~\cite{truong2021learning} to assist in the assessment. Specifically, we first predict the dense correspondence between the input and edited images along with a confidence value for each pixel. Then we warp the edited images back to the original ones and measure their similarity.


\setlength{\arrayrulewidth}{0.5mm}
\setlength{\tabcolsep}{3pt}
\renewcommand{\arraystretch}{1.0}
\begin{table}[t]
	\centering
	\caption{Quantitative comparison of appearance consistency across different methods. 
 }  
        \small
	\label{tab:comparison_metric}
	\centering
	\begin{tabular}{ccccc}
		\toprule
		Metric     & OBJect-3DIT & Zero123 & Stable-Zero123 & \emph{Ours} \\
		\midrule
        Perceptual Similarity $\uparrow$ & 0.796 & 0.859 & 0.870 & \textbf{0.905} \\
        LPIPS $\downarrow$ & 0.137 & 0.144 & 0.132 & \textbf{0.102} \\
        Mean Confidence $\uparrow$ & 0.164 & 0.098 & 0.146 & \textbf{0.203} \\
        Confident Area $\uparrow$ & 0.279 & 0.152 & 0.245 & \textbf{0.356} \\
		\bottomrule
	\end{tabular}
\end{table}

\begin{figure*}[t]
	\centering
\includegraphics[width=0.8\textwidth]{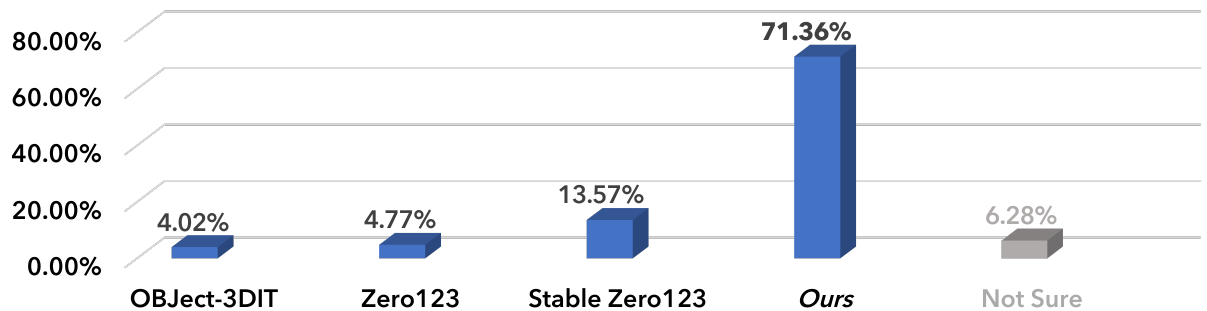}
	\caption{User preferences of the editing results from different methods.}
	\label{fig:comparison_user}
\end{figure*}

The reported appearance consistency metrics include the perceptual similarity between input and edited images, average LPIPS distance between the input and flow-warped edited images, the mean confidence and the area of confident regions predicted by the dense matching method. The perceptual similarity and LPIPS are calculated for images downsampled to $224^2$. Confident regions are determined by applying a threshold of $0.25$ to the confidence values provided by \cite{truong2021learning}. 
All these metrics except for the perceptual similarity are calculated within the foreground object regions to highlight the editing consequences.

Table~\ref{tab:comparison_metric} shows that our method consistently outperforms the other compared ones, achieving the highest scores in all four metrics. This demonstrates its advantage in maintaining a consistent appearance throughout the editing process.

\setlength{\arrayrulewidth}{0.5mm}
\setlength{\tabcolsep}{3pt}
\renewcommand{\arraystretch}{1.0}
\begin{table}[t!]
	\centering
	\caption{Ablation study on the impact of the iterative process, measured by appearance consistency after each iteration. 
 }
	\label{tab:ablation_loop}
	\centering
	\small
	\begin{tabular}{cccc}
		\toprule
		Metric     & Iteration 1 & Iteration 2 & Iteration 3 \\
		\midrule
        Perceptual Similarity $\uparrow$ & \cellcolor{red!25}0.890 & \cellcolor{yellow!25}0.895 & \cellcolor{green!25}\textbf{0.905}\\
        LPIPS $\downarrow$ & \cellcolor{red!25}0.114 & \cellcolor{yellow!25}0.108 & \cellcolor{green!25}\textbf{0.102} \\
        Mean Confidence $\uparrow$ & \cellcolor{red!25}0.141 & \cellcolor{yellow!25}0.161 & \cellcolor{green!25}\textbf{0.203} \\
        Confident Area $\uparrow$ & \cellcolor{red!25}0.235 & \cellcolor{yellow!25}0.284 & \cellcolor{green!25}\textbf{0.356} \\
		\bottomrule
	\end{tabular}
\end{table}

\paragraph{User study.}
We also conduct a user study and collect statistical data on participant preferences. 
Throughout the questionnaire, the participants were presented with the editing results of aforementioned four methods in a randomly shuffled order. They were asked to evaluate the results from three aspects: layout plausibility, image quality, and appearance consistency, and then choose the one they deemed best (see the \emph{supp. material} for the interface).

Figure~\ref{fig:comparison_user} presents the result of the user study with 33 participants. The participants preferred  our results in 71.36\% of the cases, which is more than  5$\times$ higher than the second-best performer Stable Zero123. This study further corroborates the superior visual quality and image consistency of our method.


\subsection{Ablation Study}

\begin{figure}[t]
    \centering
    \adjustbox{valign=t}{\begin{minipage}{0.69\textwidth}
        \includegraphics[width=\textwidth]{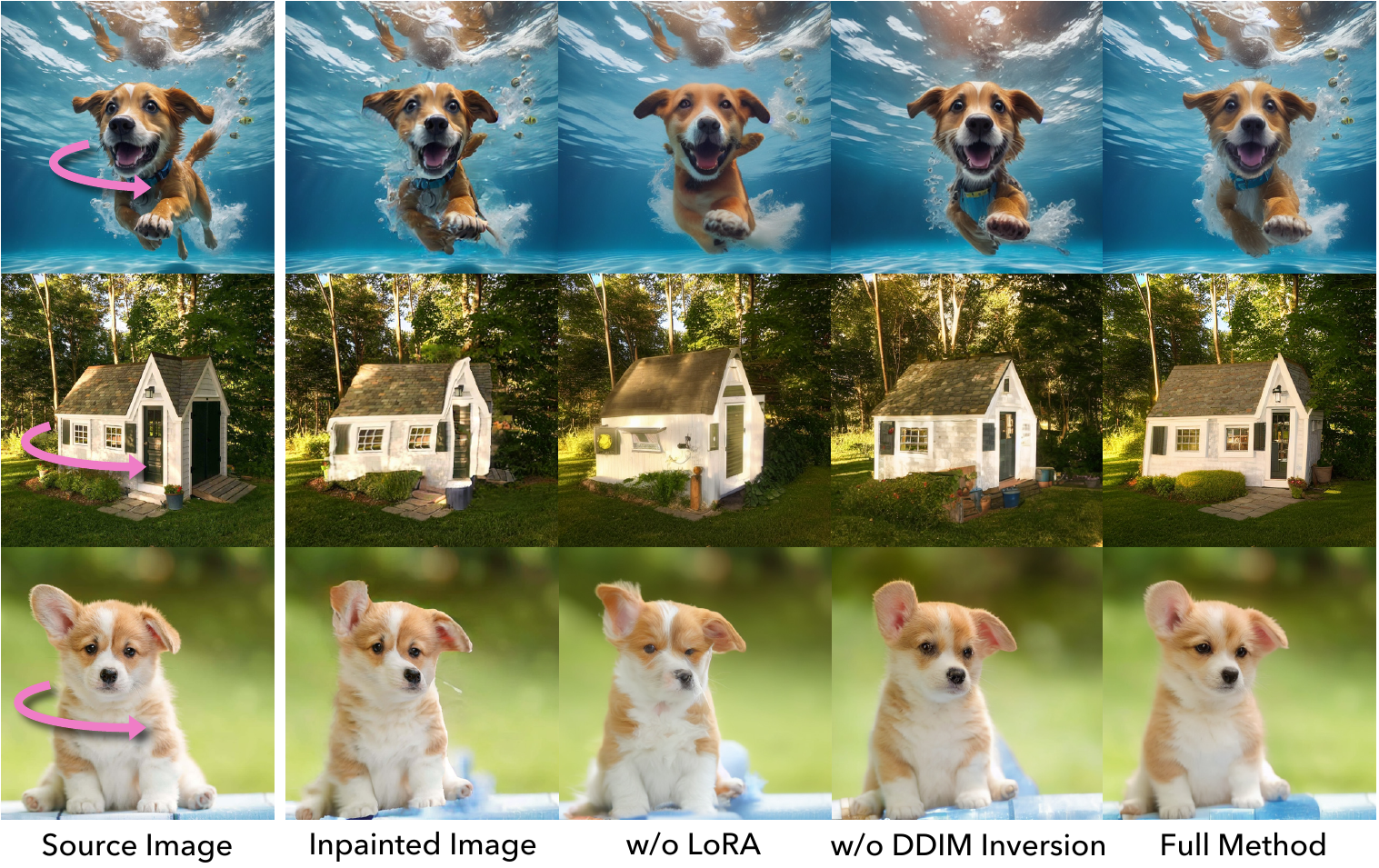}
    \end{minipage}}
    \hfill  
    \adjustbox{valign=t}{\begin{minipage}{0.275\textwidth}
        \includegraphics[width=\textwidth]{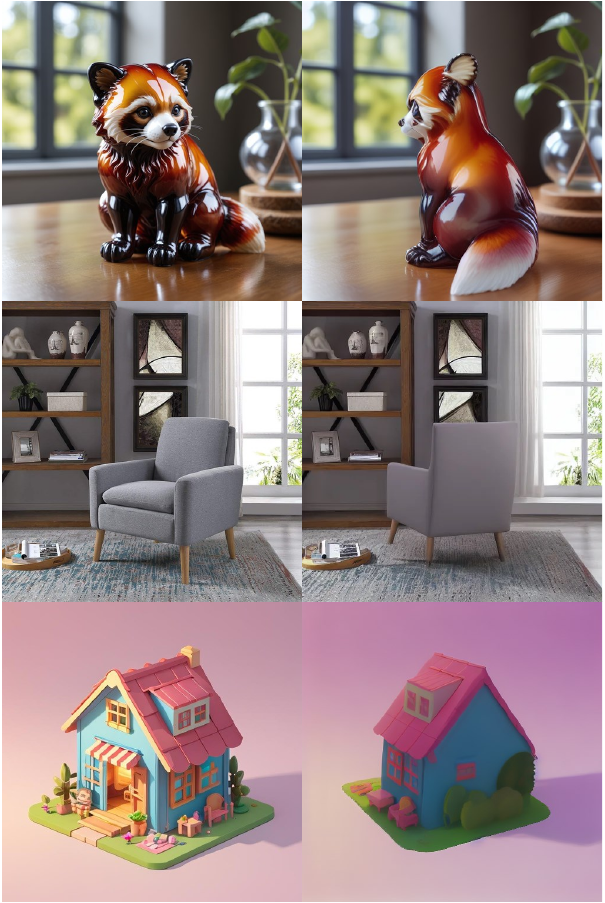}
        \vfill
    \end{minipage}} 
    \begin{minipage}[t]{0.69\textwidth}
     \caption{Ablation study on the effect of each component in the undistortion phase in the first iteration.}
     \label{fig:ablation_undistortion}
    \end{minipage}
    \hfill
    \begin{minipage}[t]{0.275\textwidth}
     \caption{Results under large transformations.}\label{fig:large_transformation}
    \end{minipage}
\end{figure}

\paragraph{Impact of undistortion designs.}
In the undistortion phase, we utilize an editing method adapted from SDEdit~\cite{meng2022sdedit} with two notable modifications: the  integration of LoRA and running forward chain with noise-free DDIM inversion. 
These enhancements are designed to retain the base diffusion model’s ability to amend distorted layouts while also maintaining appearance consistency.
We evaluate the effect of these modifications by removing them respectively under a single iteration. As shown in Figure~\ref{fig:ablation_undistortion}, omitting the LoRA negatively affects both distortion correction and the preservation of appearance consistency. Similarly, the absence of the DDIM inversion adversely affects the consistency of the image appearance. 
Combining  both modifications yields the best undistrotion results that exhibit both effective layout rectification and satisfactory appearance preservation.

\paragraph{Impact of the iterative process.}
Figure~\ref{fig:vis} and Section~\ref{sec:visual} have shown that through our iterative algorithm, the novel view synthesis and undistortion results are gradually improved upon visual inspection. Here, we provide a detailed quantitative evaluation of  appearance consistency across different iterations. 
Table~\ref{tab:ablation_loop} shows that, consistent with our visual assessment, our iterative novel view synthesis and shape alignment gradually improves the appearance consistency, demonstrating the effectiveness of our algorithm design.




\section{Limitations}

Our method still has several limitations. 
First, 
our method faces challenges in preserving the extremely fine details of the input images, a constraint tied to the capability of the pretrained base diffusion models.
Also, the physical correctness of lighting and shadow is not guaranteed, although the pretrained diffusion models possess the prior of visual harmony.
Second, under extremely large transformations, failure of inpainting models due to small visible area and significant errors in the estimated depth map can hinder the successive undistortion phase from achieving a realistic layout. To mitigate these issues, implementing small and incremental edits as can prevent significant errors, though this extends the editing duration. 
For example, Figure~\ref{fig:large_transformation} demonstrates some large transformation results created by iteratively applying smaller editing steps.
Enhancing speed and robustness in these challenging scenarios is the focus of our future work.

\section{Conclusion}

We have presented a method that advances 3D-aware single-image editing by leveraging pre-trained image diffusion models for superior open-domain image handling. Our key contribution is an innovative use of geometric priors of diffusion models with an iterative depth-assisted novel view synthesis and depth alignment process. The experimental results demonstrate that our method produces superior-quality 3D-aware edits with exceptional visual fidelity, significantly outperforming prior techniques. 
We believe our method can be a useful tool for many interesting applications in creative industries, such as graphic design, augmented reality, and video game development.


%
%
\bibliographystyle{splncs04}
\bibliography{main}

\newpage

\appendix

\begin{center}
    \textbf{
    \large  Supplementary Materials}
\end{center}

\begin{figure}[h!]
    \centering
    \includegraphics[width=\textwidth]{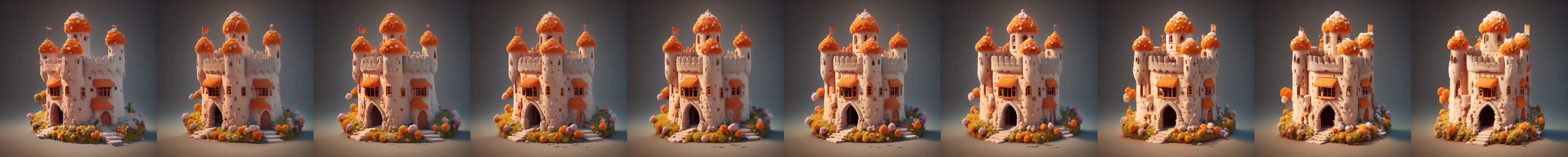}  
    \includegraphics[width=\textwidth]{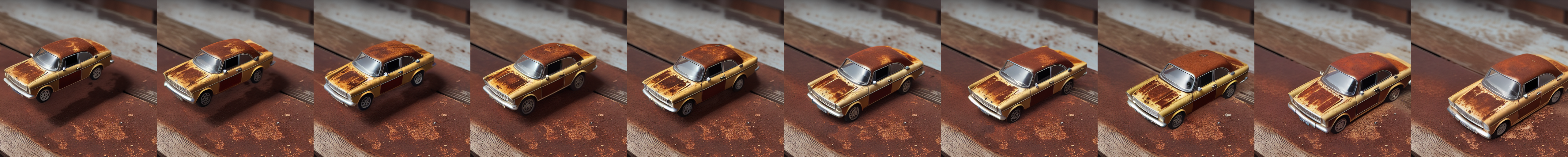} 
    \vspace{-14pt}
  \caption{Results under interpolated pose sequences. Top row: Horizontal rotation ranging from $-30^\circ$ to $30^\circ$; Bottom row: Lateral 3D translation from further left to closer right.}
  \label{fig:interpolation_sequences}  
  \vspace{-15pt}
\end{figure}

\section{More Results and Comparison}

\subsection{Results under Interpolated Pose Sequences}

For better visual examination of the shape and appearance consistency of our editing results, we present two sequences of editing results under interpolated poses in Fig~\ref{fig:interpolation_sequences}.
Specifically, we define two target transformations in oppsite directions and create an interpolated transformation sequence. Our editing method is then applied to each transformation in the sequence individually. 
As we can see, our results exhibit a high degree of shape and texture consistency under both  rotation and translation.

\begin{figure}[h!]
    \centering
    \includegraphics[width=\textwidth]{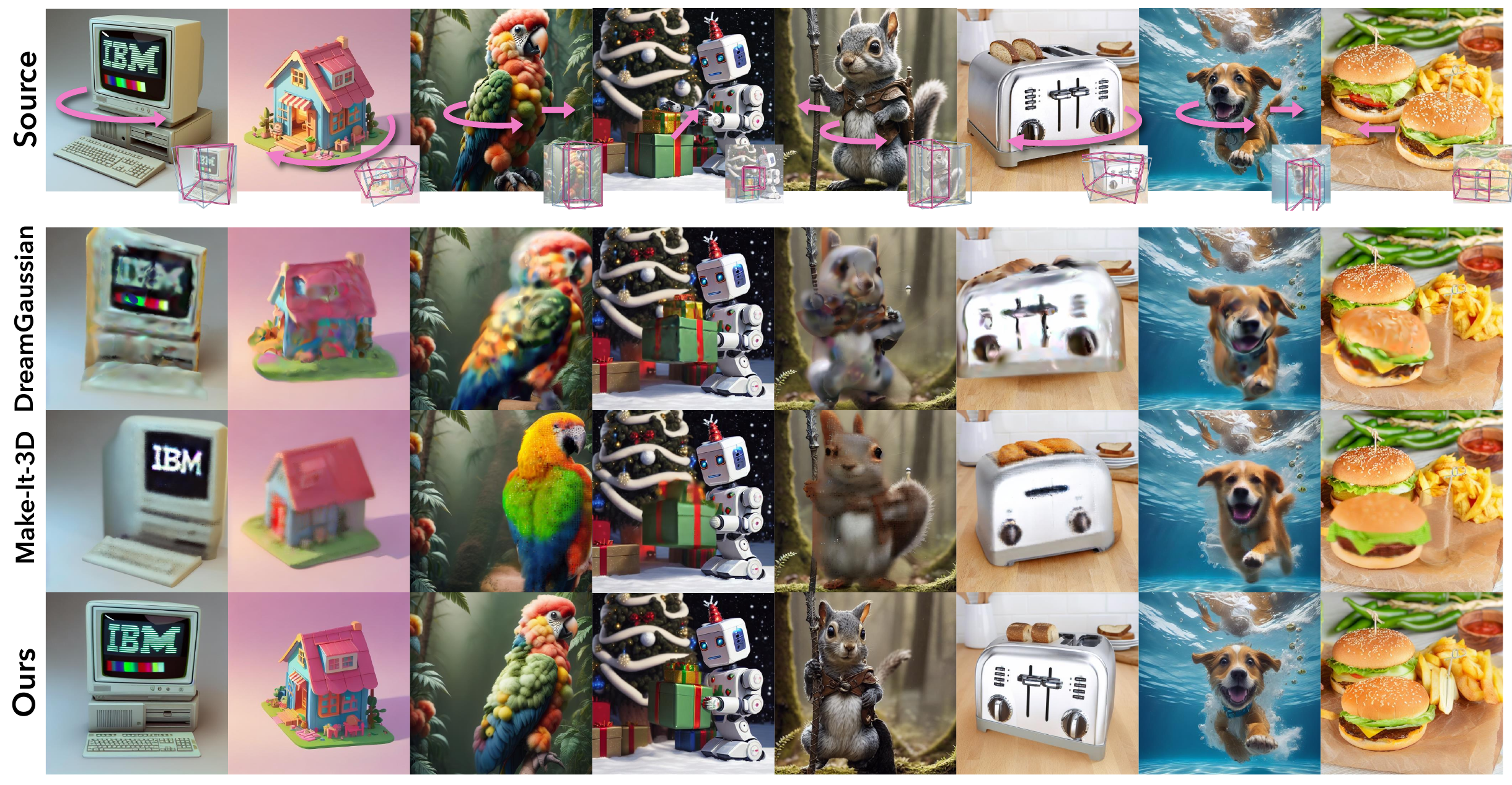}   
  \caption{Visual comparison with Make-It-3D and DreamGaussian.}
  \label{fig:comparison_imageto3d}  
\end{figure}

\subsection{Comparison with Distillation-based Image-to-3D Methods}

We additionally conducted a qualitative comparison between our approach and two image-to-3D techniques based on Score Distillation Sampling (SDS)\cite{poole2022dreamfusion}: Make-IT-3D\cite{tang2023makeit3d} and DreamGaussian~\cite{tang2023dreamgaussian}. The results are presented in Fig. \ref{fig:comparison_imageto3d}. These methods were evaluated at the $512^2$ resolution. Although they offer intact 3D geometries and hence enable $360$-degree synthesis, they suffer from apparent visual quality issues when compared to our results, such as blurriness and noticeable artifacts. Moreover, the processing time for SDS-based methods typically ranges from several minutes to hours, which is not applicable to interactive image editing scenarios that demand timely feedback.

\subsection{Results and Comparison on the Full Benchmark}

We present the editing results on the whole evaluation set, and compare the results of the four methods as in the main paper. We used
\textit{a single set of hyperparameters} as described in the implementation details for generating all our results in the benchmark. The results can be found in a seperate supplimentary document. The columns in each figure are arranged as follows: 1st column: the input image. 2nd column: the 3D edit, visualized by a source 3D bounding box (in blue and thinner) and a target 3D bound box (in red and thicker). 3rd to 6th columns: OBJect-3DIT, Zero123, Stable-Zero123, and our method.

\section{Additional Implementation Details}

\subsection{View Synthesis}

Our view synthesis phase includes two operations -- warping with depth maps and layered inpainting with base diffusion model. Here we describe the implementation in detail.

\paragraph{Depth warping.}
Starting with the source image, we convert the selected object pixels to a 3D point cloud using the specified camera intrinsics. We then create a mesh by linking points based on their original image plane positions. The warp is performed by rasterizing from the target camera's viewpoint with a fast GPU rasterizer~\cite{Laine2020diffrast}. Pixels that are excessively stretched beyond a set threshold are masked. The stretch ratio is the inverse of the smaller singular value of the Jacobian matrix between source $(u,v)$ and target $(i,j)$ image-space coordinates:
\begin{equation}
    l'=\left[\sigma_{\rm min}\left(\frac{\partial (u, v)}{\partial (i, j)}\right)\right]^{-1}.
\end{equation}

\paragraph{Layered inpainting.}
For the background inpainting in our layered inpainting approach, we use Stable Diffusion Inpaint~\cite{rombach2022high} and depth ControlNet~\cite{zhang2023adding} to generate a complete background image. The depth ControlNet helps by ensuring the depth map of the foreground is appropriately filled, taking into account planes and corners. This involves converting the depth map to a disparity map, and then to a disparity gradient map, highlighting planes as constant areas and corners as edges. We then employ an edge-flow-aware inpainting technique~\cite{bertalmio2001navier} to fill in the gradient map, followed by solving a Poisson editing~\cite{perez2023poisson} problem to complete the disparity map.

\subsection{Undistortion}

The undistortion phase involves a LoRA~\cite{hu2021lora}, DDIM inversion, and sampling~\cite{song2020denoising}. We use Stable Diffusion v1.5 from Huggingface \texttt{Diffusers} implementation throughout the undistortion phase. The conditioning text for the input image is generated by BLIP-2~\cite{li2023blip}. The LoRA is applied on \textit{to\_q}, \textit{to\_k} and \textit{to\_v} projection layers with a rank of 16 to learn the semantic features and avoid over-fitting the spatial layout. The LoRA is trained at the preparation stage with a learning rate of $5\times 10^{-4}$ and a batch size of 4 for 60 steps. For the DDIM sampling, we empirically found that a spatial mask applied to the standard deviation term of DDIM can better preserve the background region while simultaneously correct the foreground region with such spatial-adaptive random noise perturbation.

\subsection{Shape Alignment}
Our shape alignment phase including solving an optimization problem minimizing the reprojection error along with a gradient regularization term to suppress outliers. Denoting the transformation matrix as $\boldsymbol{R}$ and $\boldsymbol{T}$ and camera intrinsics matrix as $\boldsymbol{K}$, the optimization target is formulated as:
\begin{equation}
\begin{split}
    \min_{\boldsymbol{D}}&\sum_{
    (u,v)\in\boldsymbol{M}\atop
    ((u,v), (i,j))\in\boldsymbol{C}
    }\left\|\left(\frac{x_{uv}}{z_{uv}}, \frac{y_{uv}}{z_{uv}}\right)-(i,j)\right\|_2^2+\lambda\left\|\nabla\boldsymbol{D}_{uv}-\nabla\boldsymbol{D}^{(0)}_{uv}\right\|_2^2\\
    \mathrm{s.t.}&\quad\quad\quad(x_{uv}, y_{uv}, z_{uv})^{T}=\boldsymbol{K}(\boldsymbol{R}\boldsymbol{K}^{-1}(u, v, 1)^{T}\boldsymbol{D_{uv}} + \boldsymbol{T})
\end{split}.
\end{equation}
which is a sparse non-linear least-squares problem. We adopt the sparse least squares solver from SciPy~\cite{2020SciPy-NMeth} package to efficiently solve the problem. 

\subsection{Zero123 and Stable Zero123}

The Zero123 models are limited to handling object-centric images and 2-DoF rotation, including relative azimuth and relative elevation, as inputs. In contrast, our models accommodate 6-DoF transformations. To adapt to the Zero123's constraints, we preprocess the input images by removing the background and centering the object specifically for Zero123. Additionally, we meticulously decompose the transformation and compensate for the missing degrees of freedom by applying image-space translation, rolling, and scaling. To obtain the relative azimuth and elevation angles, we calculate the absolute Euler angles of gazing direction in the object space of the preprocessed object-centric image. The zoom-in and zoom-out effect is approximated by an additional image-space scaling by the fraction of center depths. In the final step, the output image is blended with the inpainted background. To avoid an unnatural look from a direct overlay, we subtly inpaint the edges within a 16-pixel perimeter for a smoother transition.

Stable Zero123 shares the same pipeline except for that it accepts the absolute elevation angle of source view as an additional input to mitigate the ambiguity of relative representation in a polar coordinate system. Therefore, we feed it with the absolute elevation angle calculated in the object space.

\begin{figure}[t!]
    \centering
    \includegraphics[width=\textwidth]{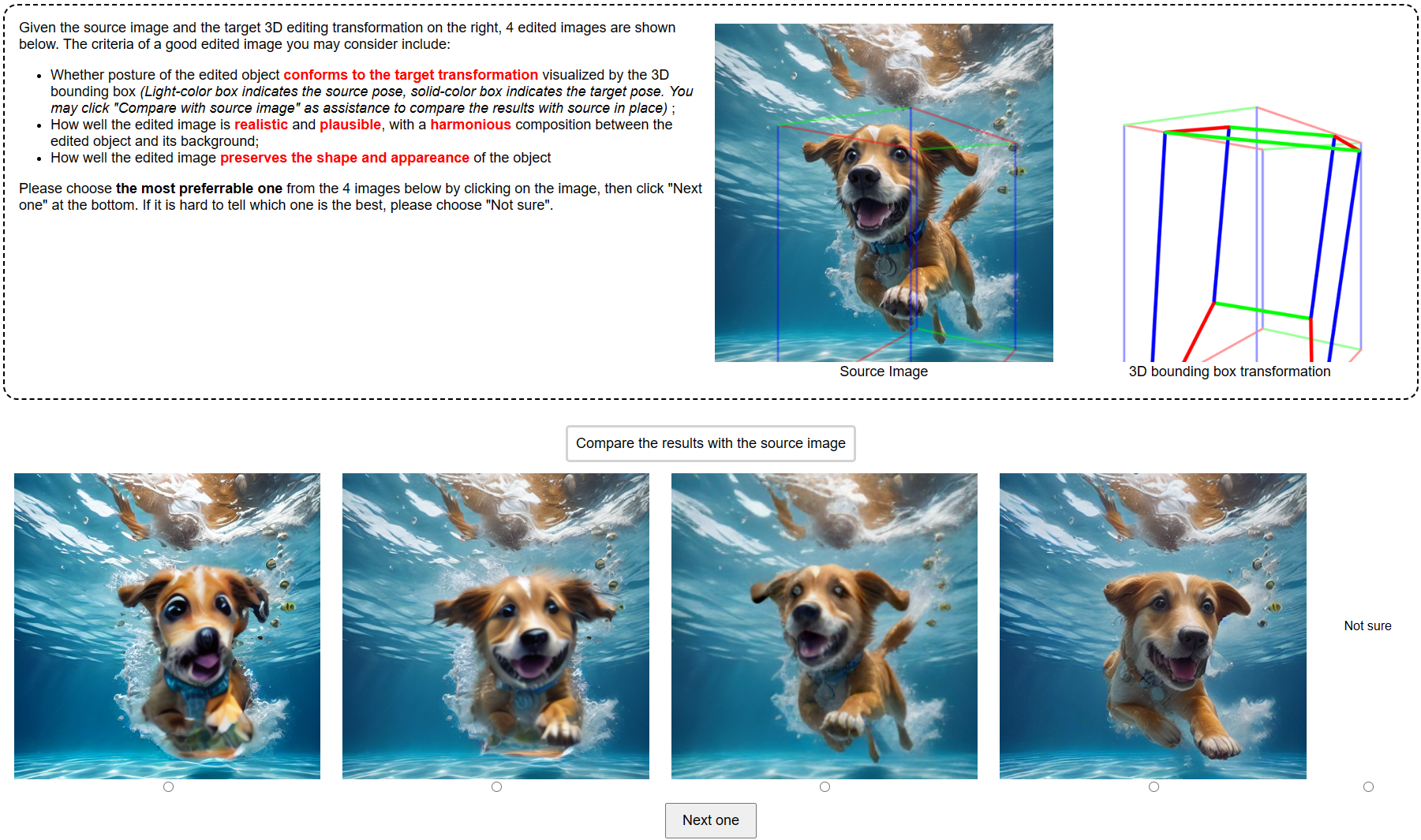}  
  \caption{Screenshot of our user study interface.}
  \label{fig:user_study_interface}  
\end{figure}

\subsection{ObJect3DIT}

OBJect3DIT has seperate models for different editing operations. We utilize their pretrained translation and rotation models and apply them one by one for combined transformation editing. The editing is instructed by  numerical values of the transformation as well as text prompts that indicate the selected object. We employ BLIP-2 to automatically generate the prompts. Specifically, we render a bounding rectangle to mark the object and prompt BLIP-2 to describe the marked object with the annotated image. Then the prompt for OBJect3DIT is constructed in the format of \emph{rotate/move the \{object description\}}.

\subsection{User Study Interface}

In our user study described in the main paper, the participants were presented with the results from four methods arreanged in a randomly shuffled order, and asked to  choose the one they deemed best. The screenshot of our user study interface is shown in Fig.~\ref{fig:user_study_interface}.




\end{document}